\documentclass[runningheads]{llncs}

\usepackage{eccv}

\usepackage{eccvabbrv}

\usepackage{graphicx}
\usepackage{booktabs}
\usepackage{siunitx}
\usepackage{enumitem}
\usepackage{nicematrix}
\usepackage[ruled,vlined,linesnumbered]{algorithm2e}

\usepackage[accsupp]{axessibility}  % Improves PDF readability for those with disabilities.

\usepackage[pagebackref,breaklinks,colorlinks,citecolor=eccvblue]{hyperref}
\usepackage{orcidlink}

\begin{document}

\title{Interpretable Perception and Reasoning for Audiovisual Geolocation} 


\author{Yiyang Su\inst{1}\orcidlink{0009-0002-4652-6828} \and
Xiaoming Liu\inst{1}\orcidlink{0000-0003-3215-8753}}

\authorrunning{Y.~Su and X.~Liu}

\institute{
Michigan State University, East Lansing, MI 48824, USA \\
\email{\{suyiyan1,liuxm\}@msu.edu}}

\maketitle

\begin{abstract}
While recent advances in Multimodal Large Language Models (MLLMs) have improved image-based localization, precise global geolocation remains a formidable challenge due to the inherent ambiguity of visual landscapes and the largely untapped potential of auditory cues. In this paper, we introduce Audiovisual Geolocation, a framework designed to resolve geographic ambiguity through interpretable perception and reasoning. We present AVG, a high-quality global-scale video benchmark for geolocation, comprising 20,000 curated clips across 1,000 distinct locations. To address the complexity of audiovisual geolocation, we propose a three-stage framework: (1) a Perception stage that utilizes a mixture-autoregressive sparse autoencoder to decompose noisy audio into semantically grounded "acoustic atoms"; (2) a Multimodal Reasoning stage that employs an MLLM finetuned via Group Relative Policy Optimization (GRPO) to synthesize these atoms with visual features; and (3) a Precision Prediction stage using Riemannian Flow Matching on the $S^2$ manifold. Our experiments demonstrate that our framework significantly outperforms unimodal baselines. These results entail that interpretable perception of the soundscape provides a critical, orthogonal signal that, when coupled with multimodal reasoning, enables high-precision global localization.
\end{abstract}
\section{Introduction}
\label{sec:intro}

\begin{figure}[t]
    \centering
    \includegraphics[width=\linewidth]{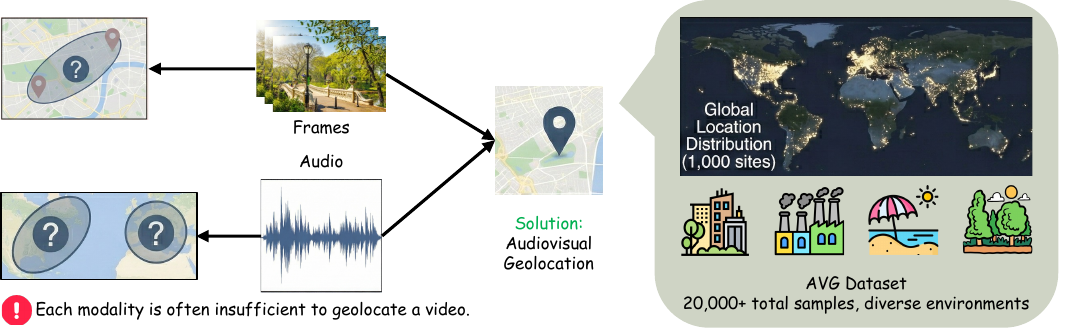}
    \caption{In the visual-only approach, ambiguous features like trees and bridges can lead to multiple location candidates. In the audio-only approach, overlapping urban sounds create complex signals that are hard to decipher. Combining modality cues, our proposed framework disambiguates candidates to pinpoint the correct location.}
    \label{fig:teaser}
\end{figure}

Geolocation, determining the geographic location of a signal or data source, remains a fundamental challenge in machine learning with profound implications for digital forensics, environmental monitoring, and autonomous navigation~\cite{bamigbade2024computer,shaamala2025machine,belmonte2019computer}. In recent years, visual geolocation~\cite{wilson2024image} has enabled applications ranging from verifying the authenticity of conflict-zone reporting to assisting search-and-rescue operations in remote environments. Similarly, text geolocation~\cite{hu2023location} has been pivotal in tracking disease outbreaks and analyzing social sentiment across specific regions. However, while video geolocation is a long-standing problem~\cite{choi2015multimodal}, the state of the art (SoTA) has largely focused on geolocation on static images. Given that only 3\% of public videos are geotagged~\cite{friedland2010cybercasing}, there is a pressing need for {interpretable perception and reasoning} frameworks that can extend these capabilities to the temporal and acoustic richness of video.

Audiovisual geolocation presents a unique challenge, as each modality may not be sufficient for geolocation on its own. As illustrated in \cref{fig:teaser}, visual geolocation may contain more information, but without integrating auditory cues, systems struggle to distinguish between visually similar environments, such as a park in London \vs a park in New York. These parks may share architecture but have vastly different acoustics. Audio is inherently noisier due to information superposition but complements visual perception. For instance, a New York park might have distant traffic hum, sirens, and subway rumble, while a London park has double-decker buses, birdsong, and church bells. A robust system must use interpretable perception to isolate sounds and multimodal reasoning to synthesize their geographic implications.

Existing geolocation models fail to adequately address the complexities of multimodal reality. Visual-only models encounter challenges in ``in-the-wild’’ scenarios where visual distinctiveness is diminished, and audio geolocation has primarily been restricted to specific, confined domains, such as natural soundscapes (\eg, bird chirps). This limitation hinders the capture of the intricate, anthropogenic sounds prevalent in human-populated environments. A significant factor contributing to this issue is the scarcity of global-scale audiovisual geolocation datasets.
This is exacerbated by a scarcity of high-quality, global-scale datasets. While user-uploaded video datasets exist~\cite{thomee2016yfcc100m}, they are often saturated with non-localizable noise—such as non-diegetic music or narrative overlays—that prevents models from learning genuine environmental correlations~\cite{astruc2024openstreetview}. 
% Recent endeavors to bridge the gap between satellite imagery and ground-level media~\cite{vyas2022gama} have attempted to fuse these two modalities. However, this approach suffers from inherent modality dissonance. The disparity between the static, outdated satellite imagery and the transient, dynamic audio further exacerbates the challenges. Additionally, the egocentric audio recording perspective introduces computational difficulties due to the nadir-oriented satellite view. 
While some other datasets~\cite{regmi2021video,lu2016geougv,friedland2010multimodal,yu2020bdd100k} provide geotagged videos, they are limited to dozens of locations, which hinders the development of generalizable audiovisual geolocation algorithms.

To address these limitations, we present the Audiovisual Geolocation (AVG) dataset, a high-quality, global-scale, audiovisual geolocation benchmark. We curate \num{12000} training, \num{4000} validation, and \num{4000} testing clips across \num{1000} distinct locations. To ensure localizability, we query public video websites for videos with location in the title or description. Additionally, we implement a rigorous filtering pipeline to ensure all samples contain diegetic audio. AVG provides the necessary foundation for training systems capable of planet-scale {multimodal reasoning}.

To tackle the challenges of audiovisual geolocation, we propose a novel three-stage framework comprising \textit{Perception}, \textit{Reasoning}, and \textit{Prediction}.

In the \textit{Perception} stage, our goal is to extract robust, discriminative features from both audio and visual streams. While visual feature extraction follows established practices, we introduce a specialized framework for the acoustic domain using a Sparse Autoencoder (SAE)~\cite{cunningham2023sparse}. The SAE is designed to decompose complex environmental soundscapes into discrete, constituent components. To ensure these components are semantically meaningful—separating, for example, a distinct bird call from background traffic noise—we pretrain the SAE on AudioSet~\cite{gemmeke2017audio}. This initialization forces the SAE kernels to learn interpretable acoustic patterns, which are then fused with visual features.

In the \textit{Reasoning} stage, we employ a Multimodal Large Language Model (MLLM) to synthesize these cross-modal signals. To adapt the general-purpose reasoning capabilities of the MLLM to the specific constraints of geolocation, we finetune the model using Group Relative Policy Optimization (GRPO)~\cite{shao2024deepseekmath}. This optimization process is guided by a set of domain-specific reward functions designed to penalize ambiguity and encourage the generation of sharp, region-specific feature representations. Finally, we employ Riemannian Flow Matching~\cite{dufour2025around} to generate the final prediction, ensuring that the geometric constraints of the Earth's surface are mathematically preserved during the inference process.

We evaluate our proposed framework on the newly collected AVG benchmark to demonstrate its effectiveness and analyze the interactions between modalities. Our experiments demonstrate that in comparison with existing audio-only geolocation methods, our {pretrained autoregressive sparse autoencoder} proves highly effective at capturing distinctive acoustic features, significantly outperforming baselines that rely on standard spectral representations or contrastive learning alone. Furthermore, when compared against SoTA visual geolocation methods, our results confirm that audio and visual modalities are fundamentally complementary. The performance of our multimodal approach consistently surpasses unimodal baselines, particularly in scenarios where visual cues are ambiguous or repetitive. This validates the hypothesis that the soundscape provides critical, orthogonal information necessary for precise localization.

To summarize, our main contributions are as follows:
\begin{itemize}[nosep]
    \item We propose a novel three-stage framework featuring a pretrained autoregressive sparse autoencoder for {interpretable perception} and an MLLM for {multimodal reasoning}.
    \item We introduce AVG, a high-quality global-scale audiovisual geolocation benchmark, containing \num{20000} video clips across \num{1000} distinct locations. % with 20 diverse clips per location.
    \item We demonstrate through extensive experiments that fusing audio and visual streams via a finetuned MLLM and Riemannian flow matching significantly improves geolocation accuracy over single-modality approaches.
\end{itemize}

\section{Related Work}
\label{sec:related_work}

\paragraph{Visual Geolocation.}
xisting literature primarily categorizes visual geolocation into regression, retrieval, and generative methods. Regression-based techniques~\cite{weyand2016planet,seo2018cplanet,muller2018geolocation,pramanick2022world,clark2023we} map images directly to geographic cells, while retrieval-based methods~\cite{yang2021cross,haas2024pigeon,wang2023fine,vivanco2023geoclip,xia2025fg,zhu2022transgeo} treat it as a nearest-neighbor search. More recently, generative models~\cite{dufour2025around} and MLLM-based methods~\cite{li2025recognition,jia2025georanker,li2024georeasoner,zhou2024img2loc} have leveraged reasoning to infer locations from subtle visual clues. Although video geolocation is a long-standing problem with benchmarks~\cite{choi2015multimodal}, the recent surge in methodological advances has been predominantly limited to static imagery. This visual-centric focus leaves a significant gap in exploiting the multimodal richness of video.

Furthermore, existing video geolocation datasets often struggle with scale and data quality. While larger-scale sets of user-uploaded content exist~\cite{thomee2016yfcc100m,friedland2011multimodal}, they are frequently saturated with ``non-localizable'' videos—clips containing non-diegetic music or visually generic interiors~\cite{astruc2024openstreetview}. Domain-specific benchmarks like BDD100K~\cite{yu2020bdd100k} are confined to driving scenarios and other datasets~\cite{lu2016geougv,friedland2010multimodal} are restricted to less than 20 locations, neither of which provide the global coverage necessary for planet-scale reasoning. Unlike these prior datasets, AVG provides a curated and globally distributed benchmark that specifically filters for high-alignment samples. By moving beyond purely visual imagery, we integrate an acoustic perception branch that resolves spatial aliasing in generic environments where even SoTA visual models fail.

\paragraph{Audio Geolocation.}
Audio geolocation~\cite{kumar2017audio,pokorny2019sound} is a relatively nascent field and \cite{chasmai2025audio} utilizes the spatial ranges of specific species as auxiliary information to improve the localization accuracy of natural sounds. However, this reliance on species-specific spatial priors makes such methods fundamentally unsuited for generalizing to non-natural, anthropogenic soundscapes. In complex human environments, the acoustic signature is defined by a dense mixture of mechanical, social, and environmental sounds that lack the structured spatial range profiles characteristic of biological species, requiring more robust, general-purpose acoustic feature extraction. {In contrast, we propose a self-supervised, mixture-aware perception framework.} \iffalse By utilizing a Sparse Autoencoder (SAE) pretrained via our Mixture-Autoregressive Training (MART) objective \fi We decompose complex environmental soundscapes into semantically interpretable acoustic atoms. This allows our model to localize a broad spectrum of human-centric and industrial sounds without requiring explicit taxonomic priors or auxiliary biological range data.

\paragraph{Cross-View Geolocation.}
Traditional cross-view geolocation~\cite{durgam2024cross,ye2024cross} is primarily formulated as a cross-modal retrieval task, where the goal is to match a ground-level query image against a database of overhead satellite imagery. While recent niche efforts have attempted to incorporate audio into this pipeline—such as pairing ground-level recordings with satellite views~\cite{vyas2022gama,pillai2024garet}—these approaches suffer from fundamental modality dissonance and perspective mismatch. Specifically, the static, nadir-oriented nature of satellite data is inherently misaligned with the dynamic, horizon-oriented nature of environmental audio. % Furthermore, because these modalities are often collected at different times and scales, they lack the temporal synchronization necessary to learn direct causal correlations.
\section{Methodology}

\paragraph{Problem Formulation.}
Let $\mathcal{V}$ and $\mathcal{A}$ denote the space of visual frames and synchronized audio waveforms, respectively. A video sample $S = \{\mathbf{v}_t, \mathbf{a}_t\}_{t=1}^T$ consists of a sequence of $T$ visual frames $\mathbf{v}_t \in \mathcal{V}$ and its corresponding audio stream $\mathbf{a}_t \in \mathcal{A}$. The goal of audiovisual geolocation is to learn a mapping function $f: (\mathcal{V}, \mathcal{A}) \to \mathbf{y}$, where the target $\mathbf{y} = (\phi, \lambda)$ represents the geographic coordinate pair (latitude and longitude) on the Earth's spherical manifold.

\paragraph{Overview.}
Our framework employs a modular three-stage design: Perception, Reasoning, and Prediction. In the Perception stage, we discover interpretable visual and acoustic features. The Reasoning stage utilizes a fine-tuned MLLM with GRPO~\cite{shao2024deepseekmath} to reason over these features and generate region-specific features. Finally, the Prediction stage generates a continuous probability density on the sphere.

\begin{figure}[tbp]
    \centering
    \includegraphics[width=\linewidth]{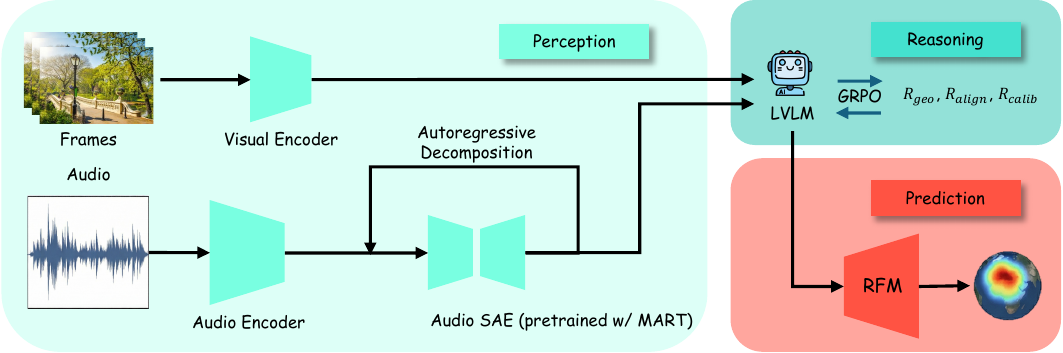}
    \caption{We process audiovisual input in three steps. (1) Perception: visual and audio encoders that extract interpretable elements from foundation models. (2) Reasoning: an MLLM, fine-tuned via GRPO, analyzes these attributes to generate geographically-rich embeddings. (3) Prediction: a Riemann Flow Matching model generate a probability density function on the Earth's surface conditioned on the reasoning output.}
    \label{fig:pipeline}
\end{figure}

\subsection{Perception: Multimodal Feature Extraction}
The perception stage aims to extract discriminative features from both visual and acoustic streams. These features serve as the grounding tokens for the subsequent reasoning stage.

% \begin{figure}[t]
%     \centering
%     \includegraphics[width=\linewidth]{figs/audio_sae.png}
%     \caption{Supervised concept pretraining with class-specific sub-dictionaries.}
%     \label{fig:audio_sae}
% \end{figure}

\paragraph{Audio Perception via IC-SAE.}
Environmental audio is inherently a mixture of overlapping signals. To extract features that are both discriminative for geolocation and semantically interpretable, we propose an audio perception branch that augments a frozen audio foundation model with an {Iterative Convolutional Sparse Autoencoder (IC-SAE)}.

% \paragraph{Architecture and Dictionary Partitioning.}
Let $f_{\text{audio}}(\cdot)$ be a frozen CLAP~\cite{wu2023large} encoder that maps a raw audio waveform $\mathbf{a}$ to a global embedding $\mathbf{x} \in \mathbb{R}^d$. While $\mathbf{x}$ captures a high-level summary, it often collapses the diverse acoustic ``atoms'' present in a complex scene. We introduce a dictionary $\mathbf{W}_d \in \mathbb{R}^{d \times N}$, where $N=4096$ is the number of latent kernels. To ensure semantic interpretability, we partition $\mathbf{W}_d$ into $C=527$ reserved "blocks" corresponding to AudioSet~\cite{gemmeke2017audio} classes. Each class $c$ is allocated a sub-dictionary $\mathbf{W}_{d,c} \approx 8$ kernels to capture intra-class variance (\eg, varying engine displacements in the "car" class).

\begin{figure}[t]
    \centering
    \includegraphics[width=\linewidth]{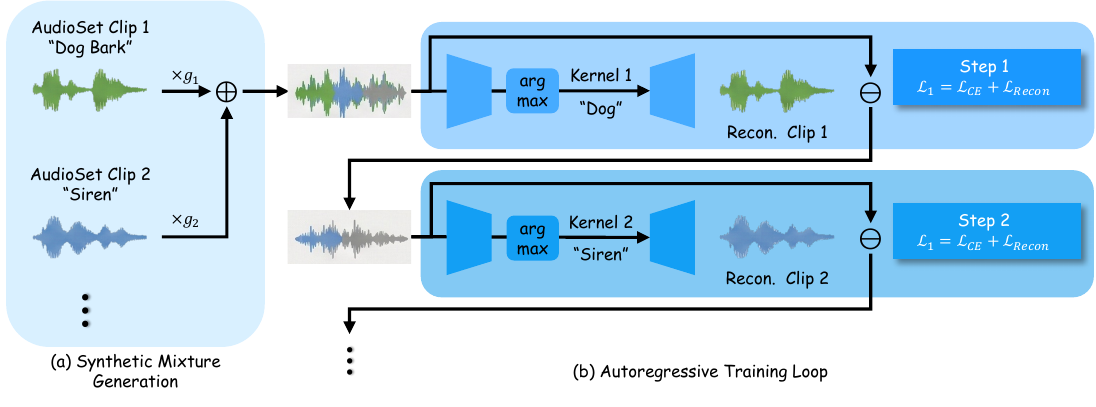}
    \caption{(a) We randomly sample clips from AudioSet to generate synthetic mixtures by a weighted sum, where the weights are randomly assigned but monotonically decreasing gains ($g_1 > g_2 > \dots$). (b) We employ an autoregressive pipeline to iteratively decompose the audio. In each iteration, we select the kernel from the dictionary with the highest activation and subtract the reconstructed audio from the mixture. The cross-entropy loss  ensures the correct semantic labels and reconstruction loss minimizes the difference between the reconstructed audio and the original audio.}
    \label{fig:pretraining}
\end{figure}

\paragraph{Mixture-Autoregressive Training (MART).}
To train the IC-SAE to handle "in-the-wild" recordings, we introduce the MART framework. MART uses a synthetic "Data Factory" to generate complex mixtures $\mathbf{x}_{\text{mix}}$ with a strictly enforced loudness hierarchy. For a set of $K$ randomly sampled clips $\{\mathbf{c}_k\}_{k=1}^K$ from AudioSet~\cite{gemmeke2017audio} with assigned energy gains $g_1 > g_2 > \dots > g_K$, the input to the IC-SAE is the CLAP embedding of the mixture:
\begin{equation}
    \mathbf{x}_{\text{mix}} = f_{\text{audio}}\left(\sum_{k=1}^K g_k \mathbf{c}_k\right).
\end{equation}
The model then recovers the sequence of classes $\mathbf{Y}_{\text{seq}} = [(\text{class}_1), \dots, (\text{class}_K)]$ step-by-step.

\begin{algorithm}[tbp]
\SetAlgoLined
\DontPrintSemicolon
\SetKwInOut{Input}{Input}\SetKwInOut{Output}{Output}

\Input{Audio mixture embedding $\mathbf{x}_{\text{mix}}$, dictionary $\mathbf{W}_d = [\mathbf{w}_{d,1}, \dots, \mathbf{w}_{d,N}]$, number of iterations $K$, ground truth targets $\{(\text{class}_k, \mathbf{c}_k, g_k)\}_{k=1}^K$}
\Output{Reconstruction residuals $\mathbf{r}_K$, selected kernels $\{j^*_k\}_{k=1}^K$}

\BlankLine
Initialize residual $\mathbf{r}_0 \leftarrow \mathbf{x}_{\text{mix}}$\;

\For{$k \leftarrow 1$ \KwTo $K$}{
    % \tcp{1. Argmax Selection Stage}
    Compute activations for all kernels: $\mathbf{a}_k \leftarrow \mathbf{W}_d^\top \mathbf{r}_{k-1}$\;
    Select the index of the most prominent kernel: $j^*_k \leftarrow \text{argmax}_{j} (\mathbf{a}_{k,j})$\;
    Predict class from selected kernel: $\text{pred\_class}_k \leftarrow \text{Class}(\mathbf{w}_{d,j^*_k})$\;
    
    \BlankLine
    % \tcp{2. Reconstruction Stage}
    Estimate gain/activation: $z_k \leftarrow \text{ReLU}(\mathbf{w}_{d,j^*_k}^\top \mathbf{r}_{k-1})$\;
    Generate reconstruction: $\hat{\mathbf{x}}_k \leftarrow z_k \mathbf{w}_{d,j^*_k}$\;
    
    \BlankLine
    % \tcp{3. Optimization}
    $\mathcal{L}_k \leftarrow \text{CrossEntropy}(\text{pred\_class}_k, \text{class}_k) + \|\hat{\mathbf{x}}_k - f_{\text{audio}}(g_k \mathbf{c}_k)\|_2^2$\;
    
    \BlankLine
    % \tcp{4. Residual Update}
    Update residual by subtracting the selected atom: $\mathbf{r}_k \leftarrow \mathbf{r}_{k-1} - \hat{\mathbf{x}}_k$\;
}
\Return{$\mathbf{r}_K, \{j^*_k\}_{k=1}^K$}
\caption{Iterative Audio Decomposition}
\label{alg:sae}
\end{algorithm}

\paragraph{Iterative Reconstruction Loop.}
The IC-SAE decomposes the mixture via a residual-update mechanism (\cref{alg:sae}).
% \begin{enumerate}
%     \item The model receives the current residual embedding $\mathbf{r}_{k-1}$ (where $\mathbf{r}_0 = \mathbf{x}_{\text{mix}}$).
%     \item A selector module $S(\mathbf{r}_{k-1})$ predicts the activation $z_{k,j}$ for the $j$-th kernel in the dictionary. Due to the class-partitioning of $\mathbf{W}_d$, this prediction implicitly identifies the most prominent AudioSet class present in the residual.
%     \item The decoder generates a reconstruction $\hat{\mathbf{x}}_k = \sum_{j} z_{k,j} \mathbf{w}_{d,j}$, where $\mathbf{w}_{d,j}$ is the $j$-th column of $\mathbf{W}_d$.
%     \item The model is optimized using a composite loss:
%     \begin{equation}
%         \mathcal{L}_k = \text{CrossEntropy}(\text{pred\_class}_k, \text{class}_k) + \|\hat{\mathbf{x}}_k - f_{\text{audio}}(g_k \mathbf{c}_k)\|_2^2
%     \end{equation}
%     \item The residual is updated for the next step: $\mathbf{r}_k = \mathbf{r}_{k-1} - \hat{\mathbf{x}}_k$.
% \end{enumerate}
By iteratively ``subtracting'' the most prominent sound embeddings, the IC-SAE isolates secondary acoustic cues that are often the most geographically distinctive, providing a set of sparse, interpretable features $\mathbf{z}$ for the subsequent reasoning stage.

\paragraph{Visual Perception.}
Given a video sequence $S = \{\mathbf{v}_t, \mathbf{a}_t\}_{t=1}^T$, we process the visual frames $\mathbf{v}_t$ using a SoTA visual geolocation backbone (\eg, GeoCLIP~\cite{vivanco2023geoclip}) pretrained on massive sets of geo-tagged imagery. To handle the temporal dimension of the video, we extract the latent feature vector $\mathbf{h}_t \in \mathbb{R}^{d_v}$ for each sampled frame and compute the temporal average:
$
    \mathbf{\bar{h}}_v = \frac{1}{T} \sum_{t=1}^T \mathbf{h}_t
$
This global visual descriptor $\mathbf{\bar{h}}_v$ captures the stationary geographic markers—such as architecture, vegetation, and infrastructure—present throughout the clip.

\subsection{Reasoning: GRPO-Finetuned MLLM}
The reasoning stage acts as the bridge between raw perception and coordinate prediction. We employ an MLLM to synthesize the visual features $\mathbf{v}_t$ and the sparse acoustic atoms $\mathbf{z}_t$ extracted by the IC-SAE. To align the model's high-level reasoning with the geographic reality of the Earth's surface, we finetune the MLLM using GRPO guided by three specialized reward functions.

\paragraph{Hierarchical S2 Geometry Reward ($R_{\text{geo}}$).} 
To avoid the "Paris problem" (linguistic ambiguity) and the "border problem" (political vs. environmental dissonance), we discard text-based semantic labels in favor of hierarchical geospatial indexing. We utilize the S2 geometry library to represent the Earth as a hierarchy of nested cells. For a predicted coordinate $\mathbf{\hat{y}}$ and ground truth $\mathbf{y}$, the reward is defined as:
\begin{equation}
    R_{\text{geo}} = \sum_{l \in \{L_1, L_5, L_{12}\}} w_l \cdot \mathbb{I}(\text{Cell}_l(\mathbf{\hat{y}}) = \text{Cell}_l(\mathbf{y})),
\end{equation}
where $w_l$ are level-specific weights and $\mathbb{I}$ is the indicator function. This rewards the model for landing in the correct geographic tile at increasing resolutions, effectively capturing environmental similarity without being penalized by arbitrary political boundaries.

\paragraph{Entity-Consistency Reward ($R_{\text{align}}$).} 
To mitigate hallucinations where the model's reasoning trace contradicts its coordinate prediction, we implement an entity-consistency check. Instead of embedding noisy reasoning strings, we use a lightweight parser to extract location entities $E_{\text{trace}}$ (\eg, "Amazon", "Sahara") from the model's output. The reward is defined as:
\begin{equation}
    R_{\text{align}} = \max_{e \in E_{\text{trace}}} (\text{Contains}(\text{Polygon}(e), \mathbf{\hat{y}})),
\end{equation}
where $\text{Polygon}(e)$ is the geographic boundary of the extracted entity. This provides a hard consistency constraint; if the model reasons about ``Canada'' but predicts a point in the ``USA'', the reward is zero.

\paragraph{Uncertainty Calibration Reward ($R_{\text{calib}}$).} 
In data-scarce or ambiguous regions (\eg, a generic tropical beach), the model must represent uncertainty rather than making a high-confidence guess. Since our prediction stage utilizes a flow-based distribution, we utilize the Negative Log-Likelihood (NLL) of the ground truth under the predicted distribution $P_\theta$:
\begin{equation}
    R_{\text{calib}} = \log P_\theta (\mathbf{y} \mid \mathbf{v}, \mathbf{a}).
\end{equation}
By maximizing this likelihood, the model is incentivized to produce sharp distributions for recognizable landmarks and diffuse distributions for ambiguous soundscapes, effectively solving the "donut problem" of multimodal geographic distributions.

\subsection{Prediction: Riemannian Flow Matching on $S^2$}

Our framework maps reasoning features from the MLLM to a precise geographic coordinate $\mathbf{y} \in S^2$, the spherical manifold of the Earth.

To avoid distortions and singularities in Euclidean regression, we use {Riemannian Flow Matching (RFM)}~\cite{dufour2025around}. We define a probability path $p_t(\mathbf{y})$ that transforms a base distribution at $t=0$ to the target distribution at $t=1$. % The flow is governed by a time-dependent vector field $\mathbf{u}_t(\mathbf{y})$ on the tangent space $T_{\mathbf{y}} S^2$.

Let $\psi$ be the reasoned embedding output by the MLLM. We learn a network $v_\theta(\mathbf{y}, t, \psi)$ to approximate the target vector field. The training objective is:
\begin{equation}
    \mathcal{L}_{\text{RFM}} = \mathbb{E}_{t, \mathbf{y}_1, \mathbf{\epsilon}} \left[ \| v_\theta(\mathbf{y}_t, t, \psi) - \dot{\mathbf{y}}_t \|^2 \right],
\end{equation}
where $\mathbf{y}_t$ is the point on the geodesic connecting the noise $\mathbf{y}_0$ and the truth $\mathbf{y}_1$ at time $t$. RFM ensures mathematical consistency with Earth’s geometry using exponential and logarithmic maps of the $S^2$ manifold. During inference, we sample from this flow to obtain a point estimate or an uncertainty-aware heatmap of potential locations.

\section{The Audiovisual Geolocation (AVG) Dataset}

\subsection{Data Collection Pipeline}

To facilitate research in synchronized multimodal localization, we introduce the Audiovisual Geolocation (AVG) benchmark. Unlike existing datasets that rely on static imagery or temporally disjointed audio-satellite pairs, AVG is designed to capture the dynamic, aligned relationship between what is seen and heard in the real world.
The construction of the AVG dataset involves a multi-stage pipeline that maximize geographic diversity while ensuring audiovisual alignment.

\paragraph{Location Sampling and Querying.} We begin by sampling geographic locations from the GeoNames database, focusing on a globally representative distribution. For each sampled coordinate, we query public multimedia repositories for high-resolution videos whose locations are mentioned in the titles or descriptions.

\paragraph{Temporal Segmentation and Filtering.} The retrieved raw videos are segmented into shorter, manageable clips of 10--20 seconds. To maintain the integrity of the ``audiovisual'' task, we implement a rigorous filtering stage. We discard any clips where the audio and visual streams are mismatched, such as videos with non-diegetic background music, voice-over narrations, or significant wind clipping that occludes the environmental soundscape. This ensures that the model must rely on the intrinsic relationship between the visual environment and its corresponding acoustic signature.

\subsection{Dataset Statistics and Splits}
The resulting AVG dataset comprises 1,000 distinct locations with 20 diverse clips per location, ensuring that models learn to generalize across different temporal conditions (\eg, weather, time of day) within the same locale. The dataset is partitioned into a training set of 12,000 samples and validation and test sets of 4,000 samples each. The average length of each video clip is approximately 10 seconds. The splits are location-disjoint, meaning no location in the validation or test sets appears during training, thereby enforcing a strict evaluation of the model's ability to generalize to unseen geographic regions.
\section{Experiments}

% In this section, we evaluate the performance of our proposed framework on the AVG dataset and provide a comparative analysis against unimodal and multimodal baselines. Furthermore, we demonstrate the generalizability of our IC-SAE perception branch on the iNatSounds audio geolocation dataset.

\subsection{Implementation Details}
Our framework is implemented in PyTorch. For the \textit{Perception} stage, the visual backbone is a frozen GeoCLIP-ViT-L/14, while the audio backbone is a frozen CLAP-L/14. The IC-SAE dictionary consists of $N=4096$ kernels, with MART pretrained on AudioSet for 5 epochs using an AdamW optimizer ($lr=1e-4$). 

For the \textit{Reasoning} stage, we utilize a LLaVA-v1.5-7B as our MLLM base. The GRPO finetuning is performed with a group size of $G=8$, with the KL-divergence penalty coefficient set to $\beta=0.01$. The \textit{Prediction} stage uses a 4-layer MLP to parameterize the vector field $v_\theta$ for Riemannian Flow Matching, trained for 50 epochs on the AVG training set. All experiments were conducted on 8 NVIDIA H100 GPUs.

\subsection{Main Results on AVG Dataset}
We evaluate our proposed method against several SoTA baselines on the AVG benchmark in four spatial scales: City (25km), Region (200km), Country (750km), and Continent (2500km). The baselines categorized for comparison include visual-only models such as GeoCLIP~\cite{vivanco2023geoclip}, Hybrid~\cite{astruc2024openstreetview}, GLOBE~\cite{li2025recognition}, and RFM $\mathcal{S}_2$~\cite{dufour2025around}. For visual-only models, we apply methods and average the results across all frames. In the audio-only domain, we compare against GeoCLAP~\cite{khanal2023learning}, TaxaBind~\cite{sastry2025taxabind}, and the audio variant of RFM $\mathcal{S}_2$~\cite{dufour2025around}. Finally, our integrated audiovisual approach is measured against a standard Late Fusion strategy that averages the results of GLOBE and GeoCLAP to determine the efficacy of our proposed method.

\begin{table}[t]
\centering\small
\caption{Comparison of geolocation performance on the AVG benchmark.}
\label{tab:avg_results}
\begin{NiceTabular}{@{}l *{4}{wc{0.7in}}}
\toprule
{Method} & \Block{1-1}{$\uparrow$ City \\ 25km} & \Block{1-1}{$\uparrow$ Region 200km} & \Block{1-1}{$\uparrow$ Country \\ 750km} & \Block{1-1}{$\uparrow$ Continent 2500km} \\ \midrule
\Block[fill=gray!20]{1-5}{Visual-only} \\
GeoCLIP~\cite{vivanco2023geoclip} & 6.8 & 11.0 & 20.8 & 32.7 \\
Hybrid~\cite{astruc2024openstreetview} & 5.7 & 9.8 & 18.7 & 25.6 \\
RFM \(\mathcal{S}_2\)~\cite{dufour2025around} & 6.2 & 10.9 & 19.6 & 31.0 \\ 
GLOBE~\cite{li2025recognition} & 6.8 & 10.6 & 20.5 & 32.1 \\ \midrule
\Block[fill=gray!20]{1-5}{Audio-only} \\
GeoCLAP~\cite{khanal2023learning} & 0.1 & 1.9 & 6.8 & 12.3 \\
TaxaBind~\cite{sastry2025taxabind} & 0.1 & 0.8 & 3.6 & 06.4 \\
RFM \(\mathcal{S}_2\)~\cite{dufour2025around} & 0.1 & 4.3 & 7.9 & 14.5  \\ 
Ours (audio-only) & \textbf{5.2} & \textbf{10.5} & \textbf{14.1} & \textbf{21.8} \\ \midrule
\Block[fill=gray!20]{1-5}{Audiovisual} \\
Late Fusion  & 6.3 & 10.2 & 19.3 & 30.5 \\
\textbf{Ours} & \textbf{8.3} & \textbf{12.5} & \textbf{22.8} & \textbf{35.4} \\ \bottomrule
\end{NiceTabular}
\end{table}

The results presented in \cref{tab:avg_results} reveal several key insights regarding the role of different modalities in geolocation. Most notably, our method consistently outperforms all competitive baselines in both the audio-only and audiovisual settings. In the audio-only category, our model achieves a city-level accuracy of 5.2\%, which represents a dramatic improvement over the 0.1\% achieved by previous SoTA methods like GeoCLAP~\cite{khanal2023learning} and RFM $\mathcal{S}_2$~\cite{dufour2025around}. This suggests our architecture is significantly more adept at extracting localized geographic cues from ambient soundscapes than previous attempts.

Across all evaluated methods, the data confirms that audio remains inherently less informative than visual data for precise geolocation. For instance, while our audio-only model performs impressively compared to its direct competitors, its city-level accuracy of 5.2\% still trails behind the visual-only GeoCLIP~\cite{vivanco2023geoclip} at 6.8\%. This reinforces the conclusion that visual landmarks and infrastructure remain the primary drivers for high-precision localization, even when acoustic models are highly optimized.

The most significant finding is the power of our audiovisual synergy. Our full audiovisual method outperforms every single-modality baseline, including the top-performing visual-only models. With a city-level accuracy of 8.3\% and a continent-level accuracy of 35.4\%, our approach surpasses the best visual-only model (GeoCLIP~\cite{vivanco2023geoclip}) by 1.5 and 2.7 percentage points, respectively. This demonstrates that audio provides critical complementary information that visual data alone lacks, and our specific fusion strategy effectively leverages this relationship to push the state of the art on the AVG benchmark.

\subsection{Audio Geolocation on iNatSounds}
To rigorously evaluate our proposed \textit{Perception} and \textit{Prediction} components independently of the visual stream, we conduct extensive experiments on the {iNatSounds} dataset~\cite{chasmai2025audio}. This benchmark focuses on the geolocation of natural soundscapes, presenting a significant challenge due to the high acoustic similarity across diverse geographic regions.

% \paragraph{Baselines.} 
We compare against three primary classes of SoTA audio geolocation methods: GeoCLAP, TaxaBind, and RFM $\mathcal{S}_2$. GeoCLAP~\cite{khanal2023learning} serves as a baseline by utilizing a frozen CLAP backbone followed by a standard MLP regression head for global coordinate prediction. TaxaBind~\cite{sastry2025taxabind} is a specialized multimodal model trained on ecological data. Lastly, RFM $\mathcal{S}_2$~\cite{dufour2025around} is a recent generative baseline that utilizes Riemannian Flow Matching on the sphere but relies on standard global audio embeddings, without our IC-SAE decomposition.

\begin{table}[t]
\centering\small
\caption{Audio-only geolocation performance on the iNatSounds~\cite{chasmai2025audio} dataset.}
\label{tab:inatsounds}
\begin{NiceTabular}{l *{5}{wc{0.7in}}}
\toprule
{Method} & \Block{1-1}{$\downarrow$ Median \\ Error (km)} & \Block{1-1}{$\uparrow$ City \\ 25km} & \Block{1-1}{$\uparrow$ Region \\ 200km} & \Block{1-1}{$\uparrow$ Country \\ 750km} & \Block{1-1}{$\uparrow$ Continent \\ 2500km} \\ \midrule
GeoCLAP~\cite{khanal2023learning} & 6856 & 00.2 & 01.3 & 07.0 & 24.9 \\
TaxaBind~\cite{sastry2025taxabind} & 4944 & 00.4 & 02.2 & 11.9 & 35.3 \\
RFM $\mathcal{S}_2$~\cite{dufour2025around} & 5798 & 01.4 & 05.5 & 16.8 & 40.4 \\
\textbf{Ours} & \textbf{1355} & \textbf{04.2}           & \textbf{13.8} & \textbf{34.4} & \textbf{64.6} \\ \bottomrule
\end{NiceTabular}
\end{table}

As shown in Tab.~\ref{tab:inatsounds}, our IC-SAE based approach achieves a significant reduction in median error. 
%The most striking trend is the drastic reduction in median error. 
Previous SoTA methods like TaxaBind exhibit a median error of 4,944 km, whereas our framework reduces this to {1,355 km}—a 72.6\% reduction in localization error. Furthermore, at the "Country" scale (750km), our method correctly localizes 34.4\% of natural soundscapes, more than tripling the performance of TaxaBind (11.9\%).
This performance jump entails the superiority of our MART and the class-partitioned dictionary $\mathbf{W}_d$. Unlike GeoCLAP, which relies on a global embedding that collapses distinct acoustic features, our IC-SAE successfully decomposes the signal into semantically interpretable kernels. The ability to outperform TaxaBind—which uses explicit taxonomic priors—suggests that our self-supervised MART pretraining on AudioSet discovers more geographically discriminative acoustic features than those provided by manual biological taxonomies.

\begin{table}[t]
\centering\small
\caption{Probabilistic audio geolocation performance on the iNatSounds~\cite{chasmai2025audio} dataset.}
\label{tab:inatsounds_prob}
\begin{NiceTabular}{l *{5}{wc{0.7in}}}
\toprule
{Method} & \Block{1-1}{$\downarrow$ NLL} & \Block{1-1}{$\uparrow$ Precision} & \Block{1-1}{$\uparrow$ Recall} & \Block{1-1}{$\uparrow$ Density} & \Block{1-1}{$\uparrow$ Coverage} \\ \midrule
RFM $\mathcal{S}_2$~\cite{dufour2025around} & -0.80 & 0.88 & 0.93 & 0.81 & 0.52 \\
\textbf{Ours} & \textbf{-0.96} & {0.88}           & \textbf{0.94} & \textbf{0.83} & \textbf{0.57} \\ \bottomrule
\end{NiceTabular}
\end{table}

\paragraph{Probabilistic Geolocation.}
A key advantage of our framework is its ability to model uncertainty in ambiguous regions. We evaluate the probabilistic quality of our predictions using Negative Log-Likelihood (NLL)~\cite{dufour2025around} and geographic coverage metrics. As shown in \cref{tab:inatsounds_prob}, our model achieves a superior NLL of $-0.96$ outperforming the RFM $\mathcal{S}_2$ baseline ($-0.80$). We also observe a marked increase in \textit{Coverage} ($0.57$ vs. $0.52$). These numbers entail that our {Uncertainty Calibration Reward} during the reasoning stage, combined with Riemannian Flow Matching on the $S^2$ manifold, produces a better-calibrated spatial distribution. Instead of collapsing to a single incorrect point, our model generates a diffuse but accurate probability density in ambiguous regions, effectively solving the "overconfidence" issues inherent in traditional regression models.

\subsection{Ablation Studies}

\begin{table*}[t]
\centering\small
\caption{Ablation study on the AVG test set. \textbf{P} (Perception), \textbf{R} (Reasoning), \textbf{Pr} (Prediction).}
\label{tab:unified_ablation}
\begin{NiceTabular}{@{}ccccccc *{4}{wc{0.7in}}}
\toprule
\Block{2-1}{ID} & \Block{1-3}{Stages} & & & \Block{1-3}{Rewards} & & & \Block{2-1}{$\uparrow$ City \\ 25km} & \Block{2-1}{$\uparrow$ Region \\ 200km} & \Block{2-1}{$\uparrow$ Country \\ 750km} & \Block{2-1}{$\uparrow$ Continent \\ 2500km} \\ \cmidrule(lr){2-4} \cmidrule(lr){5-7}
& Pr & P & R & $R_{\text{geo}}$ & $R_{\text{align}}$ & $R_{\text{calib}}$ & & & \\ \midrule
0 & \checkmark & & & & & & 6.2 & 10.9 & 19.6 & 31.0 \\
1 & \checkmark & \checkmark & & & & & 6.9 & 11.3 & 20.3 & 32.4 \\
2 & \checkmark & \checkmark & \checkmark & & & & 7.2 & 11.8 & 21.4 & 33.9 \\ \midrule
4 & \checkmark & \checkmark & \checkmark & \checkmark & & & 7.8 & 11.7 & 21.8 & 34.2 \\
5 & \checkmark & \checkmark & \checkmark & \checkmark & \checkmark & & 8.2 & 12.3 & 22.5 & 35.2 \\
6 & \checkmark & \checkmark & \checkmark & \checkmark & \checkmark & \checkmark & 8.3 & 12.5 & 22.8 & 35.4 \\ \bottomrule
\end{NiceTabular}
\end{table*}

\paragraph{Efficacy of MART Pretraining.} To evaluate the specific contribution of MART to audio perception, we compare it against a standard SAE baseline on the audio-only iNatSounds dataset. While a standard SAE achieves a median error of 3,920~km and a region-level (200km) accuracy of 9.5\%, the inclusion of the MART objective significantly improves these metrics to an region-level (200km) accuracy of {13.8\%}.
These results entail that the ability to ``subtract’’ prominent acoustic signals to uncover subtle, geographically discriminative secondary sounds is not an inherent property of sparse coding, but a specific capability learned through our proposed autoregressive training. This demonstrates that interpretable perception is enhanced when the dictionary is trained to decompose mixtures into their constituent semantic atoms, which is essential for localization in complex environmental soundscapes.

To evaluate the technical fidelity of the interpretable perception stage, we investigate how effectively the IC-SAE maintains its semantic grounding throughout the training process. Specifically, we evaluate the decomposition accuracy using 200 randomly generated audio mixtures, each synthesized from discrete AudioSet~\cite{gemmeke2017audio} clips with known ground-truth labels. After the initial MART phase, the IC-SAE achieves a decomposition accuracy of 92\%, successfully isolating the correct acoustic atoms from the mixtures. However, following the subsequent finetuning on the geolocation task, this accuracy drops to 68\%.

This performance shift prioritizes geographic discriminability over pure signal reconstruction, sacrificing some semantic consistency. The drop to 68\% suggests feature drift during finetuning, where acoustic kernels optimize for subtle regional variances at the expense of AudioSet category alignment. Despite this, the framework remains effective, with semantic grounding providing interpretable acoustic features for resolving spatial aliasing. This experiment confirms MART’s strong initialization for interpretable perception, but highlights a potential area for future work in preserving semantic labels during finetuning.

\paragraph{Impact of Framework Stages.} We utilize the first three rows of Tab.~\ref{tab:unified_ablation} to analyze the cumulative gains from each stage of our framework. Starting from a baseline that utilizes visual cues with a standard prediction head (ID 0), we observe that the introduction of the interpretable perception stage (ID 1) improves accuracy across all scales, notably increasing city-level accuracy from 6.2\% to 6.9\%. The subsequent integration of the multimodal reasoning stage (ID 2) provides a further boost, reaching 7.2\% at the 25km scale. The trend in these numbers entails that while synchronized perception provides a stronger feature foundation, explicit reasoning is required to synthesize these features. The jump from ID 1 to ID 2 suggests that extracting acoustic atoms is only half the battle; the model requires the logical capacity to synthesize these atoms with visual cues to resolve spatial aliases that occur at the city and region scales.

\paragraph{Role of Rewards.} We ablate our reward functions in rows 4--6 of \cref{tab:unified_ablation}. Replacing standard distance-based losses with the \textit{Hierarchical S2 Geometry Reward} ($R_{\text{geo}}$) increases city-level (25km) accuracy to 7.8\%. Adding the {Entity-Consistency Reward} ($R_{\text{align}}$) and the {Uncertainty Calibration Reward} ($R_{\text{calib}}$) leads to our final performance of {8.3\%} at the city (25km) scale and {35.4\%} at the continent (2500km) scale.
The implications of these results are twofold. First, the success of $R_{\text{geo}}$ confirms that treating the Earth as a hierarchical geometric grid is mathematically superior to Euclidean distance for spatial reasoning. Second, the incremental improvements from $R_{\text{align}}$  and $R_{\text{calib}}$ entail that enforcing strict factuality and proper confidence modeling is critical for preventing the model from making overconfident but geographically impossible predictions. This confirms that our multi-reward strategy effectively aligns multimodal reasoning with the specific geometric and semantic constraints of global geolocation.

\subsection{Qualitative Results}

\begin{figure}[tbp]
    \centering
    \includegraphics[width=\linewidth]{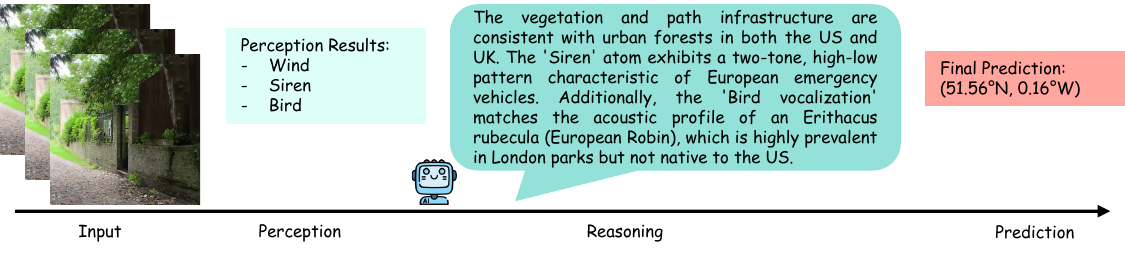}
    \caption{Qualitative result of interpretable perception and reasoning.}
    \label{fig:qualitative}
\end{figure}

To provide a qualitative understanding of the performance gains, we visualize an in-the-wild sample from the AVG test set. These examples highlight the specific scenarios where unimodal or late-fusion baselines fail, and where our framework successfully integrates cross-modal cues.
Visual cues in a temperate urban park are ambiguous, causing spatial aliasing and a split probability distribution between Europe and North America. Our perception stage resolves this by decomposing the noisy acoustic mixture into semantically grounded atoms from the AudioSet ontology. The {Wind} atom is generic, while specific {Siren} and {Bird vocalization} atoms provide discriminative evidence. The MLLM synthesizes these atoms with visual features to generate a textual reasoning chain and idenfiy the siren’s two-tone pattern and the European Robin’s call as region-specific. The final hidden state serves as the feature input for the {prediction} stage. The Riemannian Flow Matching, conditioned on the model’s logical deduction, accurately localizing the video to Hampstead Heath, London.
\section{Conclusion}

In this paper, we introduce the problem of \textit{Audiovisual Geolocation}, moving beyond the limitations of unimodal and disjointed cross-view approaches. By constructing the {AVG Dataset}—the first global-scale, natively synchronized video benchmark—we provide a foundational resource for studying the alignment of visual and acoustic signals in geographic space. Our proposed three-stage framework demonstrates that (1) sparse, semantically interpretable audio atoms extracted via our {MART-pretrained IC-SAE} are superior to global embeddings, (2) {GRPO-based reasoning} over the $S^2$ manifold effectively resolves spatial aliasing, and (3) {Riemannian Flow Matching} ensures mathematically consistent coordinate prediction. Our results consistently outperform state-of-the-art unimodal baselines, particularly in generic environments where visual cues are ambiguous. % This work establishes a new paradigm for multimodal spatial reasoning, proving that the soundscape is not merely auxiliary data, but a critical, discriminative signal for understanding our world.

% \section*{Acknowledgements}
% Please insert your acknowledgments here.

\bibliographystyle{splncs04}
\bibliography{secs/refs}
\end{document}